\documentclass{article}

\usepackage[preprint]{neurips_2022}

\usepackage[utf8]{inputenc} 
\usepackage[T1]{fontenc}    
\usepackage{hyperref}       
\usepackage{url}            
\usepackage{booktabs}       
\usepackage{amsfonts}       
\usepackage{nicefrac}       
\usepackage{microtype}      
\usepackage{xcolor}         
\usepackage{natbib}
\usepackage{array}
\usepackage{amsmath}
\usepackage{amssymb}
\usepackage{graphics}
\usepackage{graphicx}
\usepackage{subfig}

\title{Sparse Infinite Random Feature Latent Variable Modeling}

\author{%
  Michael Minyi Zhang \\
  Department of Statistics and Actuarial Science\\
  University of Hong Kong\\
  Hong Kong SAR\\
  \texttt{mzhang18@hku.hk}
}

\newcommand{\polya}{Pólya}

\newcommand{\B}{\mathbf{B}}

\newcommand{\betadist}{\text{Beta}}
\newcommand{\bigO}{\mathcal{O}}
\newcommand{\bbeta}{\boldsymbol{\beta}}
\newcommand{\bkappa}{\boldsymbol{\kappa}}

\newcommand{\bmu}{\boldsymbol{\mu}}

\newcommand{\bomega}{\boldsymbol{\omega}}
\newcommand{\bOmega}{\boldsymbol{\Omega}}
\newcommand{\bPhi}{\varphi_{w}(\X)}
\newcommand{\bZPhi}{\varphi_{w}(\X\odot\Z)}
\newcommand{\bPsi}{\boldsymbol{\Psi}}

\newcommand{\bxi}{\boldsymbol{\xi}}

\newcommand{\bSigma}{\boldsymbol{\Sigma}}
\newcommand{\bS}{\boldsymbol{S}}
\newcommand{\btau}{\boldsymbol{\tau}}
\newcommand{\btheta}{\boldsymbol{\theta}}

\newcommand{\E}{\mathbb{E}}
\newcommand{\gammadist}{\mbox{Gamma}}
\newcommand{\GP}{\mathcal{GP}}

\newcommand{\N}{\mathcal{N}}

\newcommand{\Reals}{\mathbb{R}}
\newcommand{\eye}{\mathbf{I}}
\renewcommand{\d}{\text{d}}

\newcommand{\K}{\mathbf{K}}

\newcommand{\V}{\mathbf{V}}
\newcommand{\w}{\mathbf{w}}
\newcommand{\W}{\mathbf{W}}
\newcommand{\x}{\mathbf{x}}
\newcommand{\xp}{\x^{\prime}}
\newcommand{\X}{\mathbf{X}}
\newcommand{\y}{\mathbf{y}}
\newcommand{\Y}{\mathbf{Y}}
\newcommand{\z}{\mathbf{z}}
\newcommand{\Z}{\mathbf{Z}}
\newcommand{\zero}{\mathbf{0}}

\newcommand{\U}{\mathbf{U}}

\newcommand{\Ell}{\mathcal{L}}

\begin{document}

\maketitle

\begin{abstract}
	We propose a non-linear, Bayesian non-parametric latent variable model where the latent space is assumed to be sparse and infinite dimensional \textit{a priori} using an Indian buffet process prior. \textit{A posteriori}, the number of instantiated dimensions in the latent space is guaranteed to be finite. The purpose of placing the Indian buffet process on the latent variables is to: 1.) Automatically and probabilistically select the number of latent dimensions. 2.) Impose sparsity in the latent space, where the Indian buffet process will select which elements are exactly zero. Our proposed model allows for sparse, non-linear latent variable modeling where the number of latent dimensions is selected automatically. Inference is made tractable using the random Fourier approximation and we can easily implement posterior inference through Markov chain Monte Carlo sampling. This approach is amenable to many observation models beyond the Gaussian setting. We demonstrate the utility of our method on a variety of synthetic, biological and text datasets and show that we can obtain superior test set performance compared to previous latent variable models.
\end{abstract}


\section{Introduction}
Latent variable modeling (LVM) is a popular statistical method for representing high-dimensional observations with a lower-dimensional structure. Popular examples of continuous latent variable modeling include methods like factor analysis, where the observed data are assumed to decompose into a linear combination of underlying factors, which are lower-dimensional random variables which partially explain the variability of the observed data, and the factor loadings, which represents the extent to which an observation expresses the latent factor. 
LVMs can be interpreted as matrix factorization techniques, where the observed data can be modeled as a linear combination of two low-rank matrices. However, these linear models are limited in their capacity to capture the underlying structure in the data. 

Non-linear extensions of latent variable models have proven to be quite powerful for modeling high-dimensional data. The Gaussian process latent variable model (GPVLM) is one notable example of a Bayesian, non-linear, latent variable model where the functional relationship between the latent space and the observed data is assumed to be Gaussian process distributed. However, in the Bayesian setting, inference is quite challenging as computation is generally tractable under Gaussian observation models. One method to make posterior inference tractable outside the Gaussian setting is to approximate the kernel function in the Gaussian process with the random Fourier feature approximation. Using the random projections from this approximation, we can compute simple Gibbs sampling updates in a Markov chain Monte Carlo (MCMC) sampler for GPLVM-based models outside of Gaussian observation models.

Moreover, selecting the number of latent dimensions to use in these models is an important variable but generally unknown \textit{a priori}. Typical strategies for selecting the latent dimensionality include cross validation, or selecting an large number of latent dimensions and shrinking the effect of the excess dimensions towards zero. These techniques are computationally demanding and ignore the underlying uncertainty of the latent dimensions.  

Additionally, the latent space can be difficult to interpret and may contain spurious latent dimensions that do not offer any additional information about the data. It can also be difficult to disentangle the what contributions a latent dimension might provide to modeling some underlying attribute in the observed data. Thus, enforcing sparsity in the latent variables results in a concise and interpretable latent space that furthermore can produce a regularization effect for superior predictive performance compared to non-sparse methods.

In this paper, we present a novel Bayesian method for sparse, non-linear, latent variable modeling. By combining a random Fourier features-based latent variable model with a beta-Bernoulli prior on the latent space, we can simultaneously \textit{automatically explore the number of latent dimensions}, \textit{learn a sparse representation of the latent variables}, and \textit{tractability perform posterior inference for Gaussian and non-Gaussian observations}. We will proceed with an introduction to previous work in latent variable modeling in Section~\ref{sec:background}. Then we will formally introduce our model and the inference procedure in Section~\ref{sec:method}. Next, we will examine the performance of our model on a wide variety of data sets in Section~\ref{sec:experiments}. Lastly, we conclude our paper with a discussion of possible future work in Section~\ref{sec:conclusion}.

\section{Background}\label{sec:background}
\subsection{Latent Variable Models}
The basic, linear latent variable model is expressed as a low-rank matrix decomposition: $ \Y = \X\W $. We denote the observed data with $ \Y = [ \y_1 , \ldots , \y_N ]^{T} $, which is an $ N \times J $ matrix, for $ N $ observations and $ J $ observed dimensions. The latent variables are denoted as $ \X = [ \x_1, \ldots , \x_N ]^{T} $ which is an $ N \times D $ matrix, for $ D \ll J $ latent dimensions. We denote $ \W = [ \w_1 , \ldots \w_J]$ to be a $ D \times J $ projection matrix. If we set $ \X $ to be the product of the left singular vectors and the diagonal matrix of singular values, $ \U\bSigma $ and $ \W $ to be the right singular vectors corresponding to the $ D $ largest eigenvalues of $ \Y\Y^{T} $ , then we obtain the optimal solution to principal components analysis \citep{pearson1901liii}.

We can place probabilistic assumptions on the data generating process for the linear LVM to obtain a Bayesian formulation of the basic factor analysis model \citep{press1989bayesian}:
\begin{equation}
\y_i \sim \N_{J}(\x_i\W, \sigma^{2} \eye ),\; \x_i\sim \N_{D}(0,\eye),\; \w_j \sim \N_{D}(\bmu_{\w}, \bSigma_{\w}).
\label{eqn:fa}
\end{equation}
In this context, $ \X $ are the latent factors and $ \W $ are the factor loadings matrix. By marginalizing out the latent variables, $ \X $, and estimating $ \W $ via maximum likelihood, we obtain the probabilistic variant of principal components analysis \citep{tipping1999probabilistic}. Thus, we can see the connection between factor analysis, dimensionality reduction and low-rank matrix factorization and we can unify these methods as latent variable modeling \citep{engelhardt2010analysis}. LVMs are commonly used in applied settings, where we observe different recordings corresponding to a single datum, particularly in scientific applications where we are interested in measuring traits that are not directly unobservable. For example, \cite{conti2014bayesian} analyze the 1970 British Cohort Study data set using a Bayesian factor analysis model and discover the contribution of latent, unobservable factors like depression and anxiety play in a person's life outcomes.

Outside the Gaussian setting, we can model positive integer-valued data sets using LVMs as well. Non-negative matrix factorization can model such data sets by constraining $ \X $  and $ \W $ in the matrix factorization setting to be positive valued \citep{lee1999learning}. In the probabilistic setting, we can assume the data in the generating process is Poisson distributed using Poisson factor analysis \citep[PFA,][]{zhou2012beta}:
\begin{align}
\begin{split}
y_{ij} &\sim \mbox{Poisson}(\x_i \w_j^{T}),\; x_{id}\sim\mbox{Gamma}\left(r_d, \frac{1-p_d}{p_d}\right),\; \w_j \sim \mbox{Dirichlet}(w_{0}, \ldots , w_{0})\\
r_d &\sim\mbox{Gamma}(a_0, b_0),\; p_d \sim \mbox{Beta}(c_0\epsilon_0, c_0(1-\epsilon_0)).
\end{split}
\end{align}
We can use model admixtures in using latent Dirichlet allocation (LDA), where we can discover the underlying topical content of a corpus of documents  \citep{blei2003latent,griffiths2004finding}:
\begin{align}
\begin{split}
\y_{i} \sim \mbox{Mult}(n_i; \x_i\w_1^{T} ,  \ldots , \x_i\w_J^{T} ), \; \x_i\sim \mbox{Dirichlet}(x_{0}, \ldots , x_{0}),\; \w_j \sim \mbox{Dirichlet}(w_{0}, \ldots , w_{0}).
\end{split}
\end{align}
where $ N_i $ is the total number of words in document $ i $: $ n_i = \sum_{j=1}^{J}y_{ij} $. This basic admixture model can decompose each document into a probability distribution over topics, which represents the semantical content of the document, and model a topic as a probability distribution over the vocabulary.

\subsection{Sparse Latent Variable Models}
The basic latent variable model is a powerful tool for discovering the underlying structure in a data set. However, often times we may assume that an observation is associated with only a small subset of latent factors. The sparsity constraint in a latent variable model acts as a regularizer which enforces a concise latent representation of the data and can enhance the interpretability of the latent factors \citep{olshausen1996emergence,olshausen1996natural}. The latent variables in the dense setting are unidentifiable up to an arbitrary rotation. This poses a computational problem as posterior inference algorithms for the latent variables can easily be trapped in poor local optima. Sparsity can partially solve this indeterminacy problem in latent variable models because the imposition of sparsity will eliminate many equivalent representations of the latent variables under rotational ambiguity.

A popular sparsity-inducing prior in Bayesian methods is the ``spike-and-slab'' prior that selects which latent variables are exactly zero and which latent variables are activated \citep{mitchell1988bayesian,west2003bayesian}:
\begin{equation}
x_{id} \sim (1-\pi_d) \cdot \delta_{0} + \pi_d \cdot \N(0,1),
\end{equation}
where the parameter $ \pi_d \in (0,1)$ controls the sparsity level of the latent dimension $ d $.

\subsection{Beta-Bernoulli and Indian Buffet Processes}
One drawback in the aforementioned latent variable models is that the number of latent dimensions, $ D $ is assumed to be known and fixed \textit{a priori}. In actuality, this quantity is rarely known and typical techniques for selecting the latent dimensionality like cross validation or reversible jump MCMC are extraordinarily expensive to implement in practice. A more elegant approach for selecting the latent dimensionality is to use Bayesian non-parametric (BNP) methods. The BNP perspective for this problem is to assume the number of latent dimensions is infinite \textit{a priori}. This assumption is realistic for latent variable models as we assume the number of latent dimensions grows to infinity as we observe an infinite number of observations. 

One BNP variant of spike-and-slab variable selection is using the beta-Bernoulli process, which is a distribution over infinite dimensional binary matrices. Suppose $ \Z = [\z_1, \ldots \z_N]^{T} $ is an $ N \times D $ binary-valued matrix and let the elements of $ \Z $, $ z_{id} $, be distributed as:
\begin{equation}
z_{id} | \pi_d \sim \mbox{Bernoulli}(\pi_d),\; \pi_d \sim \mbox{Beta}(\alpha/D, 1).
\end{equation}
By letting $ D \rightarrow \infty $ we obtain the beta-Bernoulli process and by integrating out $ \pi_d $ we obtain the Indian buffet process (IBP) which results in the following stochastic process for generating binary matrices \citep{thibaux2007hierarchical,griffiths2011indian}: The first observation occupies a $ \mbox{Poisson}(\alpha) $ number of latent dimensions. Each subsequent observation $ i=2, \ldots , N $ selects the previously instantiated latent dimensions with probability proportional to its popularity and then samples $ \mbox{Poisson}(\alpha / i) $ new dimensions. $ \alpha $ has plays a similar role to $ \eta  $ in the Dirichlet process as the concentration parameter for a beta-Bernoulli process, where this parameter controls the number of new features to sample.

\cite{knowles2011nonparametric} uses the Indian buffet process for sparse latent factor modeling:
\begin{align}
\begin{split}
\Z\sim\mathcal{IBP}(\alpha),\; \x_{i} \sim \N_{D}(0,\eye), \w_j \sim \N_{D}( \bmu_{\w}, \bSigma_{\w})\\
\alpha\sim\mbox{Gamma}(\alpha_0, \beta_0),\; \y_i \sim \N_{J}((\x_i\odot\z_i )\W, \sigma^{2} \eye ).
\end{split}
\end{align}
This sparse factor model is a Bayesian non-parametric variant of the factor analysis model in Equation~\ref{eqn:fa} with an infinite dimensional spike-and-slab prior to control the sparsity of the latent space.

\subsection{Non-linear Latent Variable Models and Gaussian Processes}
The linear latent variable model is limited in its expressivity and cannot capture non-linear relationships between the latent variables and the observed data. One common prior over real-valued functions to model non-linear relationships is the Gaussian process \citep{williams2006gaussian}: $ f \sim \mathcal{GP}(\mu(\cdot), \K(\cdot,\cdot)) $, given a mean function $ \mu(\cdot) $ and a covariance function $ \K(\cdot,\cdot) $, which has the property that a GP-distributed function evaluated at a finite set of points is multivariate normal: $ f(\X)\sim\N(\mu(\X), \K(\x,\x^{\prime})) $.

Gaussian processes are have been widely used in non-linear latent variable modeling \citep{lawrence2005probabilistic}. The basic GPLVM model is
\begin{equation}
\y_j \sim \N_{N}(f(\X), \sigma^{2}_j \eye),\; f(\X)\sim\N(\mu(\X), \K(\x,\x^{\prime})),\; \x_i \sim \N_{D}(0,\eye).
\end{equation}
Due to the conjugate relationship between the likelihood and prior, we can integrate out the function $ f $ in closed form--this property in crucial for computation in GPLVMs:
\begin{equation}
\y_j \sim \N_{N}(0, \K(\x,\x^{\prime}) + \sigma^{2}_j \eye),\;  \x_i \sim \N_{D}(0,\eye).
\end{equation}
The GPLVM has proven to be a very popular method for non-linear latent variable modeling. However, posterior inference for the GPLVM is computationally tractable when the likelihood is Gaussian. For non-Gaussian likelihoods, posterior inference becomes difficult because the mapping function $ f $ cannot be marginalized out in closed form. Without marginalization, we cannot evaluate closed form expressions of the posterior gradient with respect to the latent variables and therefore cannot use sampling algorithms like Hamiltonian Monte Carlo \citep{duane1987hybrid} or optimization techniques like evaluating the \textit{maximum a posteriori} (MAP) estimate.

\subsection{Random Fourier Features and Kernel Learning}
To achieve computational tractability in general settings, we can use random Fourier features (RFF) to approximate the kernel function in GPs \citep{rahimi2008random}. Although RFFs are typically used to reduce the computational complexity of kernel methods from $ \bigO(N^{3}) $ to $ \bigO(NM^2) $, we can take advantage of the random feature mapping induced by the RFF approximated GP-model that makes computation tractable for non-Gaussian likelihoods.

According to Mercer's theorem, we can equivalently represent a positive definite kernel function as an inner product of a feature mapping \citep{mercer1909xvi}:
\begin{equation}
\K(\x,\xp) = \left\langle \varphi(\x), \varphi(\xp) \right\rangle,
\quad
\x,\xp \in \Reals^{D}.
\end{equation}
Bochner's theorem states that any continuous, stationary, kernel function, $\K(\x, \xp) = \K(\x - \xp)$, is positive definite if and only if $\K(\cdot,\cdot)$ is the Fourier transform of a non-negative measure $p(\w)$, which is guaranteed to be a density with a properly scaled kernel, $ \K(\x,\x) = \K(0) = 1$ \citep{bochner1959lectures}. Let $\varphi(\x)=\exp(i \w^{\top} (\x - \xp))$, then: 
\begin{equation}
\K(\x - \xp)
= \int_{\Reals^D} p(\w) \exp(i \w^{\top} (\x - \xp)) \d\w = \E_{p(\w)}[\varphi(\x)\varphi(\x^{\prime})^{*}].
\label{eq:rffs_mc_int}
\end{equation}
Using these two theorems, we can obtain an unbiased Monte Carlo estimate of the kernel function:
\begin{equation}
\K(\x - \xp) \approx \frac{1}{M}\sum_{m=1}^{M}\varphi(\x)\varphi(\x^{\prime})^{*},
\end{equation}
for $ M $ Monte Carlo samples. If we have a real-valued kernel, then we can drop the imaginary portion of $ \varphi(\x) $, so that
\begin{equation}
\begin{aligned}
\varphi(\x) = \frac{1}{\sqrt{M}} \left[
\sin(\w_1^{\top} \x),\; \cos(\w_1^{\top} \x) \cdots \sin(\w_{M}^{\top} \x), \cos(\w_{M}^{\top} \x)\right]^{T}
\label{eq:rff_def},
\quad
\mathbf{w}_m \sim p(\mathbf{w}). 
\end{aligned}
\end{equation}
Using Mercer's theorem, we can equivalently represent a kernel method as a linear model with respect to the basis function projection, $  \varphi(\x) \beta  $, given some regression weights $ \beta \in \Reals^{M}$. This random basis function representation is vital to make posterior inference tractable in non-Gaussian likelihoods for our method.

\subsection{Random Feature Latent Variable Modeling}
\cite{lazaro2010sparse} first used RFFs to approximate a GP-regression model, but only estimated the random frequencies, $ \W $, and kernel hyperparameters using MAP estimation. \cite{oliva2016bayesian} introduced a fully Bayesian variant of a random Fourier features, called Bayesian non-parametric kernel learning (BaNK), and additionally placed Dirichlet process (DP) mixture of Gaussian-Inverse Wishart distributions on $ \W $. 
\begin{equation}
\begin{aligned}
\y_j &\sim \N_{N} \big(  \bPhi \bbeta_j , \sigma^{2}_{j}\eye \big), \bbeta_j \sim \N_{M}(\bbeta_0, \B_0), \w_m \sim \N_{D}(\bmu_{\zeta_m}, \bSigma_{\zeta_m}),\\
(\bmu_{k}, \bSigma_{k}) &\sim \mathcal{NIW}(\bmu_0, \kappa_0, \lambda_0, \bPsi_0), \zeta_m \sim \mathcal{CRP}(\eta), \eta  \sim\mbox{Gamma}(a_\eta, b_\eta), \sigma^{-2}_{j} \sim \mbox{Gamma}(a_0,b_0).
\end{aligned}
\label{eq:bank}
\end{equation}
BaNK assigns each $\w_m$ in $\smash{\W = [\w_1 \dots \w_{M}]^{\top}}$ to a mixture component using variable $\zeta_m$, which is distributed according to a Chinese restaurant process~\citep[CRP,][]{aldous1985exchangeability}, the posterior predictive distribution of the Dirichlet process. Using this DP mixture, we can explore the posterior space of stationary kernels by sampling $ \w_m $, as a random Fourier feature variant of the spectral mixture kernel \citep{wilson2013gaussian}. The parameter $ \eta $ is the concentration parameter of the Dirichlet process, which controls the number of new mixtures the DP mixture model will allocate.

$ \bbeta $ acts as the regression weights for the random basis functions of $ \bPhi $. Placing a Gaussian prior on $ \bbeta $ allows and an inverse gamma prior on $ \sigma^{2}_j $ for $ \bbeta $ and $ \sigma^{2}_{j} $ to be marginalized out in closed form for the Gaussian model. For count data likelihoods, like the Bernoulli \citep{polson2013bayesian}, negative binomial \citep{zhou2012lognormal}, and multinomial distributions \citep{chen2013scalable}\, we can easily Gibbs sample $ \bbeta $ in closed form using the \polya-gamma augmentation.

\cite{gundersen2020latent} extended BaNK for approximating GPLVMs and showed that inference was made tractable for non-Gaussian likelihoods. The resulting random feature latent variable modeling (RFLVM) extents the BaNK model in Equation~\ref{eq:bank} with:
\begin{equation}
\begin{aligned}
\y_j &\sim \Ell \big( g\big( \bPhi \bbeta_j \big), \btheta \big),
&
\btheta &\sim p(\btheta), 
& 
\x_i &\sim \N_{D}(\zero, \eye).
\label{eq:rflvm}
\end{aligned}
\end{equation}
where $ \Ell(\cdot) $ is a likelihood function, $ g(\cdot) $ is a link function that maps the real-valued $ \bPhi \bbeta_j $ to the support of the likelihood, and $ \btheta $ are other parameters to $ \Ell $, if they exist.

\section{Method}\label{sec:method}
In our method, we extend the contributions of BaNK in Equation~\ref{eq:bank} and RFLVM in Equation~\ref{eq:rflvm} with a infinite, sparsity inducing prior on $ \X $. Specifically, we place the sparse latent factor model from \cite{knowles2011nonparametric} on the RFLVM from \cite{gundersen2020latent}:
\begin{equation}
\begin{aligned}
\y_j \sim \Ell \big( g\big( \bZPhi \bbeta_j , \btheta \big),\; \x_{i} \sim \GP(0,\eye),\;\w_m \sim \GP(\bmu_{\zeta_m}, \bSigma_{\zeta_m}),\; \zeta_m\sim\mathcal{CRP}(\eta), \\
\bmu_{k}\sim\GP\left(\bmu_0, \lambda_0^{-1}\bSigma_k \right),\; \bSigma_k^{-1}\sim\mathcal{WP}(\bPsi_0, \nu_0 ),\; \Z \sim\mathcal{IBP}(\alpha), \alpha\sim\mbox{Gamma}(\alpha_0, \beta_0).
\end{aligned}
\end{equation}
Now that the latent dimension is infinite \textit{a priori}, the $ D $-dimensional priors are all assumed to be infinite dimension. Thus, the prior on the latent variables $ \X $, the random frequencies $ \W $, and the mixture locations $ (\mu_k, \bSigma_{k}^{-1}) $ are now Gaussian process and Wishart process distributed \citep{wilson2011generalised}. 

\subsection{Posterior Inference}
We adopt a fully Bayesian approach for our proposed model and take samples from the posterior using MCMC. To sample the sparsity indicators, $ \Z $, we adopt the stick-breaking representation of the IBP \citep{teh2007stick}, where we explicitly instantiate the Beta distributed weights of the latent dimensions according to a generative process for the weights that exhibit a decreasing ordering:
\begin{equation}
\nu_{d}\sim\mbox{Beta}(\alpha,1),\; \pi_d = \prod_{d^{\prime}=1}^{d}\nu_{d^{\prime}},\; z_{id}\sim\mbox{Bernoulli}(\pi_d)
\end{equation}
This stick-breaking construction is equivalent to the IBP, but allows for an efficient slice-sampling algorithm for sampling the posterior of $ \pi_d $ as opposed to the fully collapsed sampler typically used in the IBP \citep{damlen1999gibbs}. We sample the sparsity indicator, $ z_{id} $ from the full conditional:
\begin{equation}
P(z_{id} =1 | -) \propto \frac{\pi_d}{\pi^{\ast}}P(\y_i| \w_d, \x_i, \z_{i,-d}, z_{id} =1, \pi_d, \bbeta ),\; \pi^{\ast} = \min \left\{1, \min_{d:\exists z_{id}=1} \pi_d \right\}.
\end{equation}
The $\pi^{\ast}$ term in the denominator is necessary for singleton values where changing the value of $ z_{id} $ may result in a different value of the last active dimension weight. Taking advantage of the conjugate relationship between the Bernoulli and beta distributions, we can update the $ d = 1, \ldots , D^{+} $ instantiated latent dimensions:
\begin{equation}
\pi_d\sim\mbox{Beta}\left(n_d, 1 + N -  n_d \right),\; n_d = \sum_{i=1}^{N}z_{id}
\end{equation}
To sample new dimensions, we first sample $ s \sim\mathcal{U}(0,\pi^{\ast}) $, a slice variable which controls the instantiation of the dimension weights. Then we draw the new dimension weights, $ \pi_{d^{\ast}} \in (0 , \pi_{d^{\ast}-1}) $ for $ d^{\ast} = 1,\ldots , D^{\ast} $, while $ \pi_{d^{\ast}} > s $ from the distribution:
\begin{equation}
P(\pi_{d^{\ast}}) \propto \exp\left(\alpha  \sum_{i=1}^{N} \frac{1}{i} \left( 1- \pi_{d^{\ast}} \right)^{i} \right)\pi_{d^{\ast}}^{\alpha-1}(1-\pi_{d^{\ast}})^{N},
\end{equation} 
When we instantiate new dimension weights we also instantiate the new parameters corresponding to the new dimensions, $ (\x_{d^{\ast}}, \w_{d^{\ast}}, \bmu_{d^{\ast}}, \bSigma_{d^{\ast}} )$, by sampling from their respective priors.

The full conditional for the IBP concentration parameter $ \alpha $ can be sampled in closed form, if the prior on $ \alpha $ is Gamma distributed \citep{griffiths2011indian}:
\begin{equation}
P(\alpha| \Z) \sim\mbox{Gamma}\left(\alpha_0 + D^{+}, \beta_0 + \sum_{i=1}^{N}\frac{1}{i} \right)
\end{equation}

To take posterior samples of $ \X $, we use the elliptical slice sampler (ESS) \citep{murray2010elliptical}. This is an efficient MCMC algorithm to take posterior samples of non-conjugate models with Gaussian priors. We choose to use this sampling algorithm because it is easy to implement, the is ESS guaranteed to transition to a new state every iteration, and there are no tuning parameters to set.

We take posterior draws of $\W$ using a Metropolis–Hastings sampler, where our proposal distribution is set to the prior distribution on $ \w_m $. This simplifies the acceptance probability to the ratio of likelihoods:  
\begin{equation}
\begin{aligned}
\w_m^{\star} \sim  p(\w_m\mid \zeta, \bmu, \bSigma), \;
\rho_{{MH}} = \min \Bigg\{1, \frac{p(\Y \mid \X, \w_m^{\star}, \btheta)}{p(\Y \mid \X, \w_m, \btheta)}\Bigg\}, \\
P(\w_m|-) \sim \rho_{{MH}} \cdot \delta_{\w_m^{\star}} + (1-\rho_{{MH}}) \cdot \delta_{\w_m}.
\label{eq:w_sampling}
\end{aligned}
\end{equation}
and we sample the latent indicators of the DP mixture, $\mathbf{\zeta} = [\zeta_1 \dots \zeta_{M}]$, using the standard ``Algorithm 8'' for sampling for Dirichlet process mixture models, with the mixture locations integrated out \citep{neal2000markov}:
\begin{equation}
\begin{aligned}
p(\zeta_m = k \mid \bmu, \bSigma, \W, \alpha) \sim
\begin{cases}
\frac{n_k^{-m}}{M - 1 + \alpha}  \mathcal{T}(\bmu_{k}^{\prime}, \bSigma_{k}^{\prime}, \kappa_{k}^{\prime})   & n_k^{-m} > 0
\\
\frac{\alpha}{M - 1 + \alpha} \mathcal{T}(\bmu_{0}, \bSigma_{0}, \kappa_{0}) & n_k^{-m} = 0.
\end{cases}
\end{aligned}
\label{eq:zm_sampling}
\end{equation}
where $ \mathcal{T} $ refers to the multivariate-$ t $ distribution, parameterized by the mean, $ \bmu_{k}^{\prime} $, covariance, $ \bSigma_{k}^{\prime} $, and the degrees of freedom, $ \kappa_{k}^{\prime} $:
\begin{equation}
\begin{aligned}
n_k &= \sum_{m=1}^{M}I(\zeta_m = k), \; \bar{\w}^{(k)} = \frac{1}{n_k} \sum_{m:\zeta_m=k} \w_m, 
&
\kappa_{k}^{\prime} &= \kappa_0 + n_k - D^{\ast} + 1\\
\bS_{\bar{\w}_{k}} &= \sum_{m:z_m = k} (\w_m - \bar{\w}^{(k)})(\w_m - \bar{\w}^{(k)})^{\top}, 
& 
\bS_{k} &= \sum_{m:z_m = k}\frac{\lambda_0 n_k}{\lambda_0 + n_k} (\w_m - \bmu_0)(\w_m - \bmu_0)^{\top}\\
\bSigma_{k}^{\prime} &= \bPsi_0 + \bS_{\bar{\w}_{k}} + \bS_{k},
&
\bmu_{k}^{\prime} &= \frac{\lambda_0 \bmu_0 + n_k \bar{\w}^{k}}{\lambda_0 + n_k},
\end{aligned}
\end{equation}
For the purpose of convenience, we re-instantiate the parameter locations for the mixture model on $ \w $. We can Gibbs sample the scale and location parameters using the typical inverse Wishart-Gaussian conjugacy relationship with a Gaussian likelihood. 
\begin{equation}
\begin{aligned}
\bSigma_k &\sim \mathcal{W}^{-1}(\bSigma_{k}^{\prime}, \kappa_0 + n_k),
\quad
\bmu_k \sim \N(\bmu_{k}^{\prime}, (\lambda_0 + n_k)^{-1} \bSigma_k).
\end{aligned}
\label{eq:mu_sigma_sampling}
\end{equation}
We sample the DPMM concentration parameter $\alpha$ by using an augmentation scheme to make sampling $\alpha$ conditionally conjugate with a Gamma prior~\citep{escobar1995bayesian}:
\begin{equation}
\begin{aligned}
\rho &\sim \betadist(\alpha + 1, M), \frac{\pi_{\rho}}{1-\pi_{\rho}} = \frac{a_\alpha + K^{+} - 1 }{M(b_\alpha - \log(\rho))},
\;\;K^{+} = \left| \{ k: n_k > 0 \} \right|,
\\
\alpha &\sim \pi_{\rho} \cdot \gammadist(a_\alpha + K^{+}, b_\alpha - \log(h)) + (1 - \pi_{\rho}) \cdot \gammadist(a_\alpha + K^{+} - 1, b_\alpha - \log(\rho)).
\end{aligned}
\label{eq_alpha_sampling}
\end{equation}
Sampling the regression weights $ \bbeta $ and the other likelihood specific parameters (if they exist) depend on the likelihood chosen for the data. For the Gaussian model we integrate out $ \bbeta_{j} $ and $ \sigma^{2}_j $, due to the conjugacy between the Gaussian and inverse Gamma priors with the Gaussian likelihood. For likelihoods that can be written as:
\begin{equation}
P(y_{ij}| \x_i, \beta_j, \w_j ) \propto c_{ij}
\frac{(\exp(\varphi_{w}(\x_i) \bbeta_j))^{a_{ij}}}{(1 + \exp(\varphi_{w}(\x_i) \bbeta_j))^{b_{ij}}} 
\label{eq:pg_aug_integral}
\end{equation}
which include the Bernoulli, negative binomial and multinomial likelihoods, then we can Gibbs sample $ \bbeta_j $ in closed form using the \polya-Gamma augmentation:
\begin{equation}
\begin{aligned}
\omega_{ij} &\mid \bbeta_j \sim \text{PG}( b_{ij}, \varphi_{w}(\x_i)^{\top} \bbeta_j), &  \textbf{V}_{\bomega_j} &= (\bPhi^{\top} \bOmega_j \bPhi + \B_0^{-1})^{-1},  \\
\bbeta_j &\mid \bOmega_j \sim \N(\textbf{m}_{\bomega_j}, \textbf{V}_{\bomega_j}), & \textbf{m}_{\bomega_j} &= \textbf{V}_{\bomega_j} (\bPhi^{\top} \btau_j + \B_0^{-1} \bbeta_0),
\end{aligned}
\end{equation}
where $\tau_{ij} = a_{ij} - b_{ij}/2$ and $\btau_j = [\tau_{1j} \dots \tau_{Nj}]^{\top}$. The full details of the \polya-gamma augmentation are available in the Appendix. The remaining inference step is sampling the posterior of the other likelihood specific parameters, $ \btheta $. In some special cases, we have a Gibbs sampling update for the parameter, like the dispersion parameter of a negative binomial distribution \citep{zhou2012augment}, but outside of these cases we require a tailored solution for posterior sampling.

\section{Experiments}\label{sec:experiments}
We apply our sparse, infinite RFLVM (IBP RFLVM) model on a wide variety of datasets. In these experiments we evaluate the performance of our model with competing models in terms of predictive test set log-likelihood for the real-valued data sets and test set perplexity on the count-valued data sets. Our competing models in the real-valued case are the RFLVM \citep{gundersen2020latent}, the Bayesian inducing point GPLVM \citep{damianou2016variational} implemented in \texttt{GPy} \citep{gpy2014}, and the infinite linear factor model (IBP LFM) \citep{knowles2011nonparametric}. In the count-valued case, we compare the sparse and dense Poisson, negative binomial and multinomial RFLVMs, with Poisson factor analysis \citep{zhou2012beta}, and latent Dirichlet allocation \citep{griffiths2004finding}.

In the following experiments, we run each model for $ 100 $ MCMC iterations. We set the number of random features to be $ M=50 $ and the number of inducing points for GPLVM to be equal to $ M $. The hyperpriors in the model are set to be vague. We initialize the number of latent dimensions to be two for the infinite models and we set the number of latent dimensions for the finite models to be the number of instantiated dimensions of the comparable infinite models (ex: we set the number of dimensions for Multinomial RFLVM and LDA to be the number of instantiated dimensions for IBP Multinomial RFLVM). For test set evaluation, we randomly hold out $ 20\% $ of the $ y_{ij} $ in $ \Y $ and evaluate each method over five trials.

\subsection{Synthetic Data}
We generate a synthetic data set based off the canonical ``Cambridge bars'' data set used in many Indian buffet process papers. 
In our synthetic example we assume there are four latent dimensions, corresponding to each shape. We sample the parameters of the IBP RFLVM from the prior data generating process and set the random frequencies $ W $ to be the shapes of the Cambridge data set. In this data set, we know that the true data generating process is sparse in the latent space, $ X $. The predictive log likelihood results for the held out test set data in this example show that the IBP RFLVM performs better than the RFLVM, GPLVM, and the IBP LFM if there is sparsity in the latent space of the data generating process as well as a non-linear process between the latent space and observations (see Table~\ref{table:reals}).


\subsection{Empirical Data}
We evaluate our method on a wide array of empirical data sets. The real-valued data sets we look at are the Wisconsin breast cancer data set, a protein expression data set for mice with trisomy and a control group, gene expression data from a study of ovarian cancer \citep{martoglio2002decomposition}, the Olivetti faces data set. The test set predictive log likelihood is reported Table~\ref{table:reals} for the aforementioned data sets, along with the Cambridge data set. We can see that the IBP RFLVM exhibits superior performance compared to the other competing methods in this experiment. 

One possible reason why the IBP RFLVM performs so well in these examples is that sparsity will provide some level of regularization compared to dense models and produce stronger predictive performance. Another reason is that the latent space for the data generating processes of these examples indeed exhibit sparsity, so the prior of the IBP RFLVM is more appropriate than the dense models which was evidenced in the Cambridge example. The IBP RFLVM and dense RFLVM generally perform better than the Bayesian GPLVM in terms of predictive log likelihood because the RFLVM models are exact Bayesian methods and can better capture the posterior uncertainty compared to the variational Bayesian GPLVM which will only provide a good point estimate but the mean-field posterior approximation will poorly capture the posterior uncertainty. Moreover, the expressivity of the non-linear mapping of IBP RFLVM, RFLVM and GPLVM produces better predictive performance than the linear IBP LFM, which performs the worst in this experimental setting. 

Next, we evaluate our model on a collection of integer-valued data sets. We look at the following data sets: A data set that counts the number of bicycles on different roads in Montreal, a data set counting the words said by members of the U.S. congress in speeches, a collection of spam and non-spam SMS messages, and the 20 Newsgroups data set. In Table~\ref{table:count}, we can see the test set perplexity results of the competing methods on these data sets. Except for the Congress data set, the NB IBP RLFVM produces the best perplexity results. Furthermore, the IBP variant models generally perform better than their dense counterparts and the RFLVM models generally perform better than the linear models. The reason for the superior performance could be for the aforementioned reasons. Additionally, negative binomial models can model over-dispersed counts where the variance of the data exceeds the mean, which the Poisson and multinomial models cannot do.

\subsection{Qualitative Assessment}
\begin{figure}
	\centering
	\subfloat[IBP RFLVM.]{\includegraphics[width=0.25\linewidth]{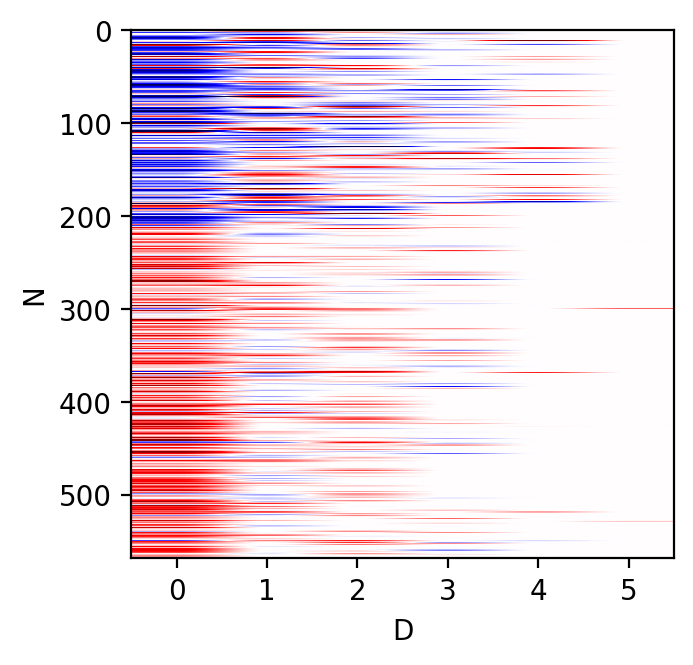}}
	\subfloat[RFLVM.]{\includegraphics[width=0.25\linewidth]{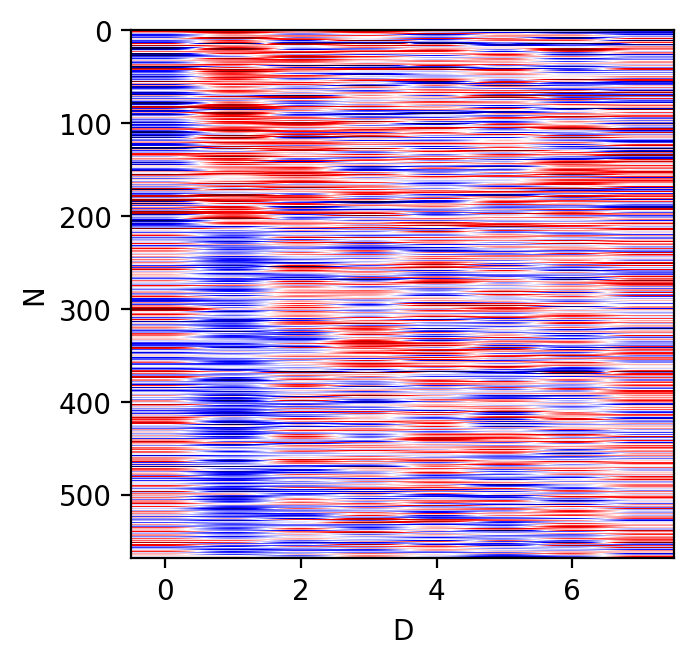}}
	\subfloat[GPLVM.]{\includegraphics[width=0.25\linewidth]{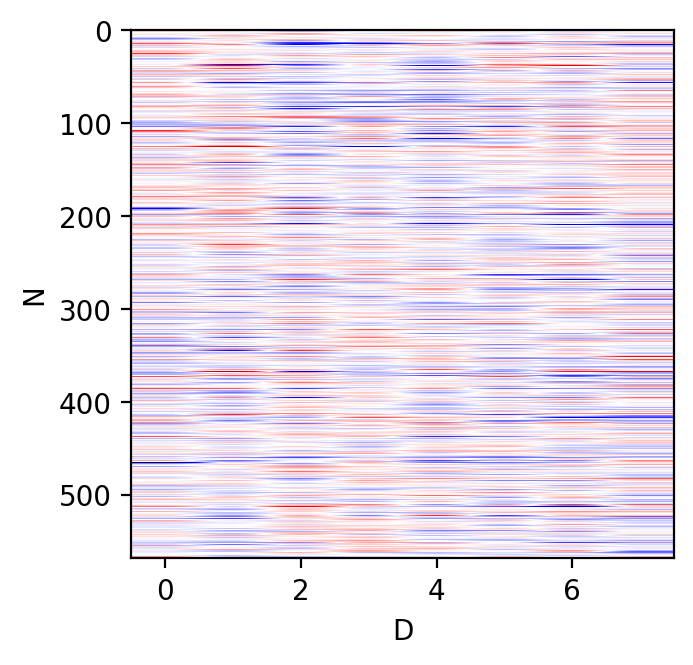}}
	\subfloat[IBP LFM.]{\includegraphics[width=0.25\linewidth]{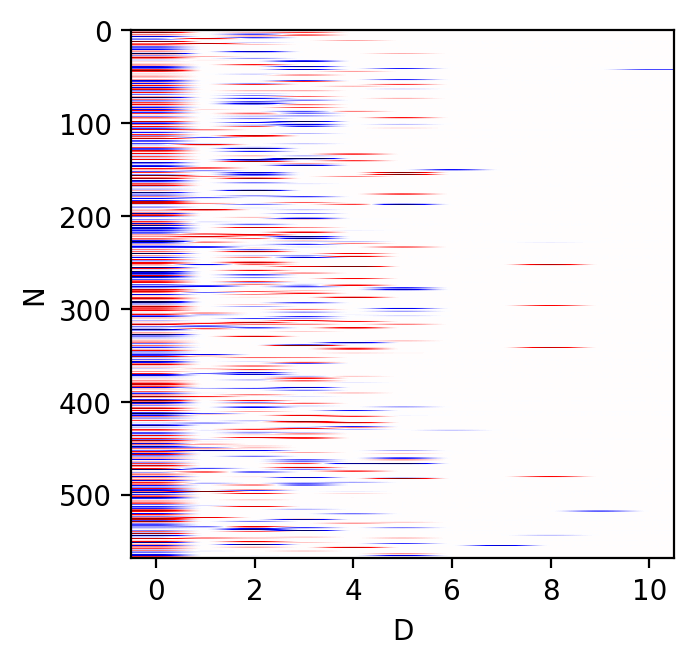}}
	\caption{Latent space for Wisconsin breast cancer data set. The first 212 observations are malignant samples, the remaining are benign.}\label{fig:cancer}
\end{figure}

\begin{figure}
	\centering
	\subfloat[NB IBP.]{\includegraphics[width=0.125\linewidth]{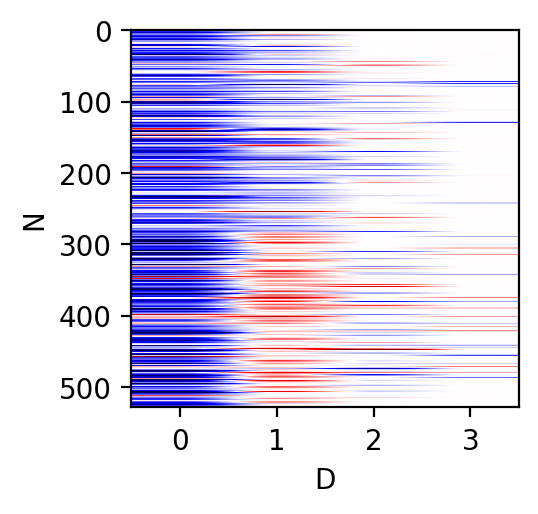}}
	\subfloat[NB.]{\includegraphics[width=0.125\linewidth]{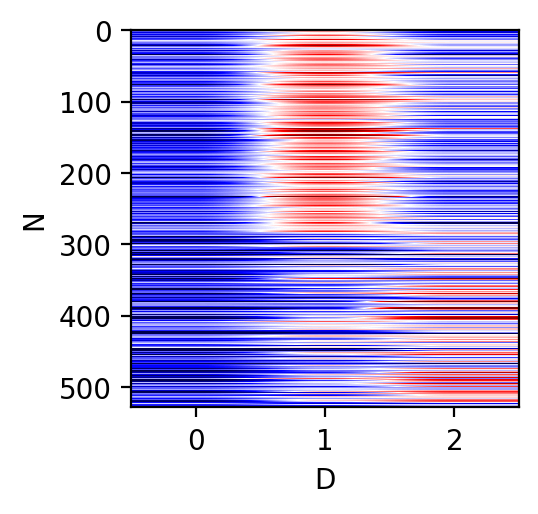}}
	\subfloat[Pois. IBP.]{\includegraphics[width=0.125\linewidth]{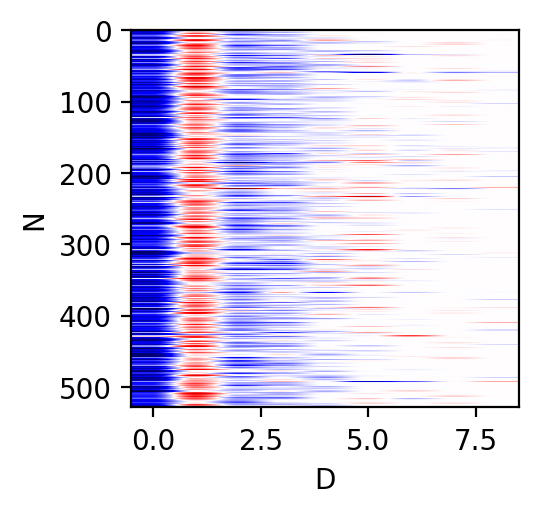}}
	\subfloat[Pois.]{\includegraphics[width=0.125\linewidth]{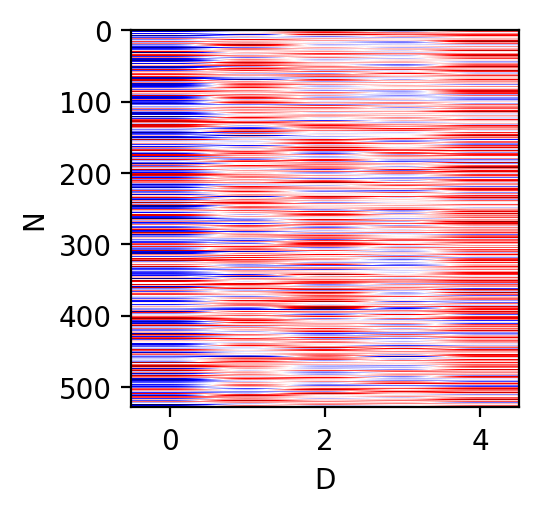}}
	\subfloat[Mult. IBP.]{\includegraphics[width=0.125\linewidth]{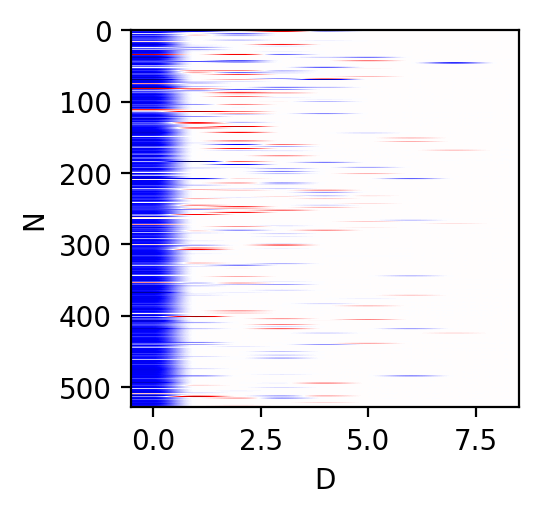}}
	\subfloat[Mult.]{\includegraphics[width=0.125\linewidth]{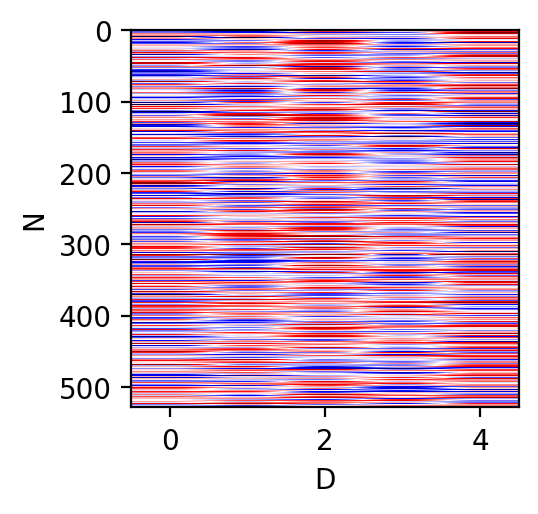}}
	\subfloat[PFA.]{\includegraphics[width=0.125\linewidth]{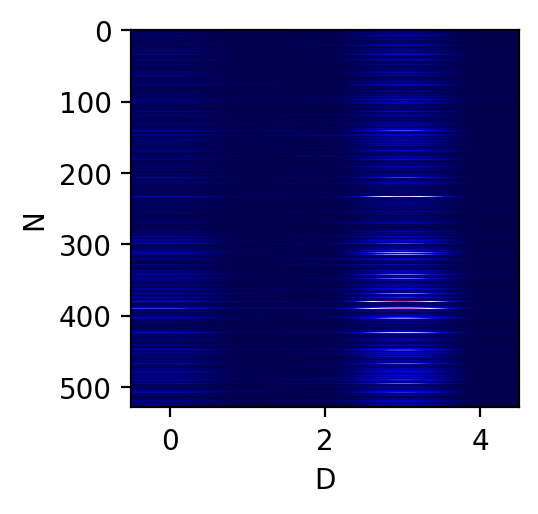}}
	\subfloat[LDA.]{\includegraphics[width=0.125\linewidth]{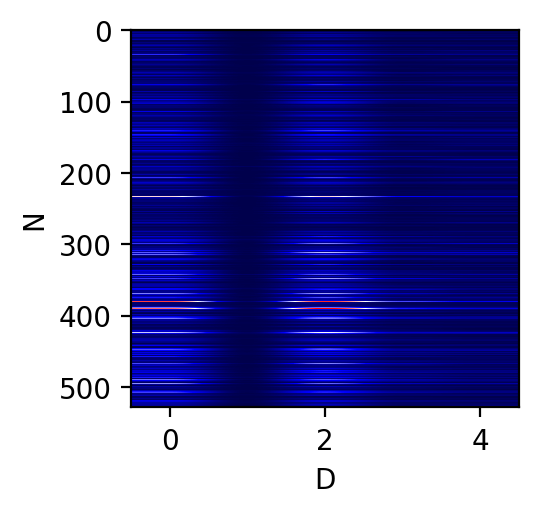}}
	\caption{Latent space for Congress data set. The first 285 observations are Democratic members of Congress, the remaining are Republicans.}\label{fig:congress}
\end{figure}

Lastly, we can look at the latent space for the Wisconsin breast cancer data set and the Congress data set and observe a clear delineation between the two classes in both of the data sets. The label information is not used in training the model, but we can uncover the two separate classes by inspecting $ \X $. Figures~\ref{fig:cancer} and \ref{fig:congress} shows the latent space of these two data sets for each of the models investigated in this experimental section. The observations in each of these figures are sorted by their label. For the IBP RFLVM and RFLVM results on the cancer data set, we can see that there is a clear delineation between cancerous and non-cancerous samples particularly with the first two latent dimensions. In the NB IBP RFLVM and NB RFLVM for the Congress data set, we can see that the second and third latent dimensions are particularly informative for discriminating between Democratic and Republican members. This structure is less obvious to see in the other models, which suggests that the best performing models in our experiments in terms of predictive performance can actually discover a clear latent structure for some of our data sets.
\begin{table}
	\centering
	\caption{Test set log likelihood for Gaussian models on real-valued datasets over five trials. One standard error is reported in parentheses.}\label{table:reals}
	\resizebox{\textwidth}{!}{
		\begin{tabular}{l|lllll}
			& Cancer & Mice & Ovarian & Olivetti & Cambridge \\ \hline
			IBP RFLVM & \textbf{6.68e+03 (114.16)} & \textbf{1.39e+04 (148.76)} & \textbf{1.81e+03 (94.02)} & \textbf{1.43e+05 (726.41)} & \textbf{1.09e+04 (1212.76)} \\
			RFLVM & 6.19e+03 (119.32) & 1.36e+04 (163.09) & 1.73e+03 (93.84) & 1.42e+05 (957.49) & 9.91e+03 (896.54) \\
			GPLVM & -4.79e+03 (26.31) & -1.22e+04 (63.53) & -6.96e+02 (92.65) & -1.16e+05 (168.80) & -2.32e+04 (134.68) \\
			IBP LFM & -2.46e+04 (33.90) & -6.10e+04 (96.61) & -3.99e+03 (21.14) & -5.84e+05 (694.63) & -2.20e+05 (6020.21)
		\end{tabular}
	}
\end{table}
\begin{table}
	\centering
	\caption{Test set perplexity for count data models on integer-valued datasets over five trials. One standard error is reported in parentheses.}\label{table:count}
	\begin{tabular}{l|llll}
		 & Montreal & Congress & Newsgroups & Spam \\ \hline
		NB IBP RFLVM & \textbf{16.25 (0.10)} & 719.05 (9.59) & \textbf{19.14 (0.27)} & \textbf{158.41 (2.20)} \\
		NB RFLVM & 16.69 (0.10) & 727.74 (6.86) & 21.18 (0.12) & 193.26 (2.42) \\
		Poisson IBP RFLVM & 24.76 (0.89) & 2029.75 (38.47) & 23.94 (0.74) & 389.70 (13.02) \\
		Poisson RFLVM & 39.34 (1.58) & 2278.67 (13.82) & 26.66 (0.74) & 410.75 (18.81) \\
		Multinomial IBP RFLVM & 21.58 (0.37) & 1005.36 (12.71) & 20.78 (0.39) & 418.42 (7.09) \\
		Multinomial RFLVM & 22.49 (0.24) & 940.92 (19.86) & 26.53 (1.04) & 370.99 (13.41) \\
		PFA & 27.11 (2.03) & \textbf{678.55 (1.94)} & 27.43 (1.08) & 186.97 (0.57) \\
		LDA & 117.03 (1.86) & 938.74 (3.44) & 19.16 (0.03) & 218.32 (0.04)		
	\end{tabular}	
\end{table}

\section{Conclusion}\label{sec:conclusion}
In this paper we propose a Bayesian model for sparse non-linear latent variable modeling that can explore the number of latent dimension \textit{a posteriori} using the Indian buffet process. Our experimental results show that our method is generally superior to the dense variant of our model and other competing linear and non-linear latent variable models. The IBP RFLVM can easily be generalized to non-Gaussian likelihoods, which is typically difficult for GPLVMs due to a lack of computational tractability in posterior inference in the non-Gaussian setting.  

In future work, we would like to explore variants of our model using the two-parameter Indian buffet process. The one-parameter IBP used in this paper controls the overall sparsity and latent dimensionality with $ \alpha $. This constraint may be too restrictive, so we may want to use the two-parameter IBP where these two behaviors are governed by their own parameters. Lastly, we are interested in using other Bayesian non-parametric priors on the sparsity indicator, $ \Z $, like the gamma-Poisson process \citep{titsias2007infinite} or the negative binomial process \citep{zhou2013negative} and compare their performance with our IBP model. Lastly, we are also interested in applying the IBP RFLVM to scientific settings. For example, GPLVMs are commonly used for modeling single cell data where the observed data are high-dimensional but we are interested in modeling a lower dimensional representation of the data. In such applications, sparse methods are appealing but there are few examples of sparse, non-linear, non-Gaussian LVMs. We fill this gap with our proposed model.

%
%

\clearpage
\bibliographystyle{apalike}
\bibliography{references}

\clearpage
\appendix

\section{Appendix}
\subsection{Experimental Details}
The Wisconsin breast cancer, the mouse trisomy, and spam data sets are from the UCI Machine Learning Repository. The Olivetti and 20 Newsgroups data sets are from the \texttt{scikit-learn} package. The Congress data set is available at \href{https://github.com/jgscott/STA380/blob/master/data/congress109.csv}{\texttt{https://github.com/jgscott/STA380/blob/master/data/congress109.csv}}. The Montreal data set is available at \href{https://www.kaggle.com/pablomonleon/montreal-bike-lanes}{\texttt{https://www.kaggle.com/pablomonleon/montreal-bike-lanes}}. The ovarian cancer data set is available at: \href{https://web.archive.org/web/20050306171817/http://www.path.cam.ac.uk/~angio/publications/martoglioetal2002/ovcadatafinal-WEB.txt}{\texttt{https://tinyurl.com/3chw6znr}}. For the real-valued data sets, we standardize the data to have zero mean and unit variance. For the count data sets, we take the square-root transformation of the data as a normalization method for variance stabalization.

\subsection{\polya-Gamma Augmentation Derivations}
Refer to Equation \ref{eq:pg_aug_integral}, which allows us to rewrite the likelihood as proportional to a Gaussian: 
\begin{equation}
P(y_{ij}| \x_i, \beta_j, \w_j ) \propto c_{ij}
\frac{(\exp(\varphi_{w}(\x_i) \bbeta_j))^{a_{ij}}}{(1 + \exp(\varphi_{w}(\x_i) \bbeta_j))^{b_{ij}}} = 2^{-b_{ij}} e^{\tau_{ij} \varphi_{w}(\x_i)_{j}} \int_{0}^{\infty} e^{- \omega \varphi_{w}(\x_i)_{j}^2 / 2} p(\omega) \text{d}\omega.
\end{equation}
A random variable $\omega$ is \polya-gamma distributed with parameters $b > 0$ and $c \in \Reals$, denoted $\omega \sim \text{PG}(b, c)$, if
\begin{equation}
\omega \stackrel{d}{=} \frac{1}{2\pi^2} \sum_{k=1}^{\infty} \frac{g_k}{(k-1/2)^2 + c^2/(4\pi^2)},
\end{equation}
where $\stackrel{d}{=}$ denotes equality in distribution and $g_k \sim \gammadist(b, 1)$ are independent gamma random variables.  We can sample $\omega$ conditioned on $\psi_{ij}$ as $p(\omega \mid \psi_{ij}) \sim \text{PG}(b_{ij}, \psi_{ij})$. If we set $a_{ij}=y_{ij}$ and $b_{ij}=1$, then we have a sampler for Bernoulli observations \citep{polson2013bayesian}. If we set $a_{ij}=y_{ij}$ and $b_{ij}=y_{ij}+r_{j}$, then we have a sampler for negative binomial observations for a dispersion parameter $r_j$ \citep{zhou2012lognormal}.

The derivation for the \polya-gamma augmentation with the multinomial likelihood depends on the reparameterization of the likelihood \citep{holmes2006bayesian,chen2013scalable}. We rewrite the likelihood as
\begin{align}
\begin{split}
p(\Y |-) &= \prod_{i=1}^{N} \frac{\Gamma \left( \sum_{j=1}^{J} y_{ij} +1 \right)}{ \prod_{j=1}^{J} \Gamma \left( y_{ij} +1 \right) } \prod_{j=1}^{J} \left(\frac{\exp \left\{  \varphi_\W(\x_i) \bbeta_j \right\}}{\sum_{j=1}^{J}\exp{\left\{ \varphi_\W(\x_i) \bbeta_j \right\}}}\right) ^{y_{ij}}\\
&\propto \prod_{i=1}^{N}\prod_{j=1}^{J} \frac{ \left( \exp \left\{ \varphi_\W(\x_i) \bbeta_j - \xi_{ij} \right\} \right)^{y_{ij}}  }{ \left( 1+ \exp \left\{  \varphi_\W(\x_i) \bbeta_j - \xi_{ij} \right\} \right)^{y_{ij}+\sum_{j=1}^{J}y_{ij}} }
\end{split}
\end{align}
Where $\xi_{ij} = \log \sum_{j^{\prime} \neq j} \exp \{ \varphi_\W(\x_i) \bbeta_{j^{\prime}}\} $. By convention and for identifiability purposes, we set $\bbeta_J = 0$. We let $\tau_{ij} = y_{ij} - \sum_{j=1}^{J}y_{ij}/2$. The likelihood is proportional to:
\begin{align}
\begin{split}
p(\Y |-) &\propto \prod_{i=1}^{N}\prod_{j=1}^{J} \exp\Big\{\tau_{ij}\Big( \varphi_\W(\x_i) \bbeta_j - \xi_{ij}  \Big)  -\frac{\omega_{nj}}{2} \Big( \varphi_\W(\x_i) \bbeta_j - \xi_{ij}  \Big)^2 \Big\}.
\end{split}
\end{align}
So this gives us a posterior w.r.t. $\bbeta_j$ as
\begin{equation}
p(\bbeta_j \mid \y_j, \X)
\propto
p(\bbeta_j) \prod_{n=1}^{N} \exp\Big\{\bkappa_{i}\Big( \varphi_\W(\x_i) \bbeta_j - \bxi_{j}  \Big)  -\frac{1}{2} \Big( \varphi_\W(\x_i) \bbeta_j - \bxi_{j}  \Big)^{T} \bOmega_n \Big( \varphi_\W(\x_i) \bbeta_j - \bxi_{j}  \Big)  \Big\}
\end{equation}
which we can rewrite into a closed form update as
\begin{equation}
\begin{aligned}
\bbeta_j \mid \bomega_j &\sim \N(\textbf{m}_{\bomega_j}, \textbf{V}_{\bomega_j})
\end{aligned}
\end{equation}
where
\begin{equation}
\begin{aligned}
\bOmega_j &= \text{diag}([\bomega_{1j}, \dots, \bomega_{Nj}])
\\
\V_{\bomega_j} &= (\bPhi^{\top} \bOmega_j \bPhi + \B_0^{-1})^{-1},
\\
\textbf{m}_{\bomega_j} &= \textbf{V}_{\bomega_j} (\bPhi^{\top} (\btau_j + \bxi_j^T \bOmega_j ) + \B_0^{-1} \bbeta_0),
\\
\btau_j &= \y_j - \frac{1}{2} \sum_{j=1}^{J}y_{ij}
\end{aligned}
\end{equation}
and we sample $\bOmega_j$ with
\begin{equation}
\begin{aligned}
\bomega_{j} \mid \bbeta_j &\sim \text{PG}\left( \sum_{j=1}^{J}y_{ij}, \bPhi \bbeta_j - \bxi_j \right)
\end{aligned}
\end{equation}

\end{document}